\documentclass[conference]{IEEEtran}
\IEEEoverridecommandlockouts

\usepackage{cite}
\usepackage{amsmath,amssymb,amsfonts}
\usepackage{algorithmic}
\usepackage{graphicx}
\usepackage{textcomp}
\usepackage{xcolor}
\usepackage{booktabs} 
\usepackage{multirow}

\def\BibTeX{{\rm B\kern-.05em{\sc i\kern-.025em b}\kern-.08em
T\kern-.1667em\lower.7ex\hbox{E}\kern-.125emX}}
\begin{document}

\title{MoFE: A Novel Mixture-of-Experts Framework with Fourier Neural Operators for Cryptocurrency Forecasting\\

\thanks{
© 2026 IEEE. Personal use of this material is permitted.
Permission from IEEE must be obtained for all other uses,
in any current or future media, including reprinting/republishing
this material for advertising or promotional purposes, creating
new collective works, for resale or redistribution to servers or
lists, or reuse of any copyrighted component of this work in other works.
DOI: 10.1109/ICBC67748.2026.11575439
}

}

\author{\IEEEauthorblockN{Bowen Liu}
\IEEEauthorblockA{\textit{School of Art and Science}\\
\textit{University of Rochester} \\
Rochester, USA \\
bliu59@u.rochester.edu}
\and
\IEEEauthorblockN{Mingming Sun}
\IEEEauthorblockA{\textit{AGI Lab} \\
\textit{Beijing Institute of Mathematical Sciences and Applications}\\
Beijing, China \\
sunmingming@bimsa.cn}
\and
}

\maketitle

\begin{abstract}
Forecasting cryptocurrency prices remains a formidable challenge due to inherent non-stationarity, abrupt regime shifts, and multi-scale stochastic dependencies. Conventional deep learning models often struggle to capture complex underlying dynamics, frequently resulting in persistent phase-lagged predictions. To address these limitations, we propose MoFE, a novel deep learning framework that integrates Fourier Neural Operators (FNOs) within a Mixture-of-Experts (MoE) architecture. Rooted in the theoretical framework of stochastic differential equations, MoFE conceptualizes cryptocurrency volatility as a superposition of multi-frequency components, which includes user network based fundamental growth, mining costs and halving mechanism caused seasonal volatility, and market sentiment-induced chaos. Specifically, specialized adaptive FNO (AFNO) and Convolution dual-domain experts learn continuous function-to-function mappings to encapsulate global spectral trends, cyclical adjustments and microstructures, while a dynamic gating based MoE mechanism enables adaptive strategy switching across diverse market regimes. Extensive experiments on Bitcoin datasets spanning January 2020 to December 2025 demonstrate that MoFE achieves state-of-the-art (SOTA) performance in both T+1 and T+5 forecasting horizons. Notably, the model effectively mitigates the phase-lag effect, delivering superior Directional Accuracy (DA) and Information Coefficient (IC). In high-fidelity simulated trading environments, these predictive gains transfer into significant excess returns and robust risk-adjusted performance, characterized by a high Sharpe ratio.
\end{abstract}


\section{Introduction}

Cryptocurrency price forecasting represents a pivotal frontier in computational finance, increasingly shifting from heuristic technical analysis toward sophisticated data-driven machine learning paradigms~\cite{hou2020pricing, amjad2017trading}. Robust predictive capabilities are fundamental to enhancing market transparency and facilitating informed decision-making for both institutional investors and regulatory stakeholders.

Unlike traditional assets anchored by fundamental yields, cryptocurrencies are governed by algorithmic scarcity and network-driven valuations—dynamics often modeled by fundamental growth guided by Metcalfe's Law, seasonal volatility driven by mining costs and halving mechanisms, and chaos induced by market sentiment~\cite{Wheatley2018Metcalfe}. The absence of intrinsic valuation anchors leaves the market susceptible to sentiment-driven volatility, frequently invalidating traditional mean-reversion frameworks~\cite{zargar2019informational}. Developing robust predictive models remains a formidable challenge due to inherent non-stationarity, abrupt regime shifts, and multi-scale stochasticity~\cite{Wheatley2018Metcalfe}. Furthermore, compared to the centuries of data in equity markets, the relatively brief history of cryptocurrencies induces a ``data scarcity" problem, particularly when analyzing low-frequency cyclical patterns like the quadrennial ``halving" events~\cite{Chan2023Halving}.

The landscape of cryptocurrency price forecasting has transitioned from heuristic technical analysis toward sophisticated data-driven machine learning paradigms. Initially, classical statistical and ensemble frameworks~\cite{tian2023bitcoin,chu2017garch,Sekhar2022XGBoost} leveraged historical inertia to model market trends. Meanwhile, some research introduced sentiment-based approaches~\cite{zargar2019informational} based on on-chain data and social media psychology to quantify market value beyond endogenous price action. However, these methods primarily capture linear dynamics or non-inherent dynamics and are often inadequate for the pervasive non-stationarity of crypto-markets.

Initially, classical statistical frameworks (e.g., ARIMA\cite{tian2023bitcoin}  and GARCH\cite{chu2017garch}) and ensemble models leveraged historical inertia to model market trends.

Currently, deep representation learning defines the state-of-the-art, utilizing non-linear optimization to decode intricate dependencies. This evolution spans Recurrent Neural Networks (RNNs) for temporal modeling~\cite{seabe2023forecasting, Dutta2020gru}, Transformers and State-Space-Models (SSMs) for long-range dependency modeling~\cite{liu2024itransformer, Sepehri2025CryptoMamba}, which would like capture cross-period stochastic volatilization. Despite their success, conventional deep learning models primarily minimize point-wise statistical errors via regression. Consequently, they often fail to internalize the underlying continuous spectral dynamics of the market, resulting in significant phase lags and an inability to adapt to rapid regime shifts.

Many existing approaches primarily optimize point-wise regression errors, which often leads to phase lag and weak adaptability to rapid market regime shifts~\cite{jamhamed2024regime, kang2025harnessing}. In addition, purely time-domain models tend to be sensitive to high-frequency market noise, which is particularly common in cryptocurrency markets~\cite{peik2025adaptive}.

Beyond traditional architectures, Fourier Neural Operators (FNOs)~\cite{Zongyi2021FNO} have redefined sequence modeling by learning continuous function-to-function mappings in the frequency domain. Notably, the Adaptive FNO (AFNO)~\cite{Guibas2022afno} optimizes this process via frequency sparsification, allowing for continuous global convolutions that are invariant to input resolution. Further advancements in frequency-domain Mixture-of-Experts (MoE)~\cite{Chen2025freqmoe} have demonstrated superior performance in modeling complex  Partial Differential Equation (PDE) systems and non-stationary time series. Motivated by the success of these spectral methods, this work introduces a frequency-domain motivated MoE-FNO framework. By modeling cryptocurrency volatility as a superposition of multi-frequency components, our framework captures the multi-scale stochasticity inherent in digital assets to improve the accuracy of  return forecasting.

The proposed MoFE framework complements these approaches by combining frequency-domain and time-domain modeling. The global FNO component captures long-range dependencies and cyclical market dynamics, while the convolutional branch extracts short-term temporal microstructures.

The main contributions of this study are as follows:
\begin{itemize}
    \item We design a novel MoE-FNO framework tailored for cryptocurrency markets, capable of modeling non-stationary volatility and abrupt regime shift through multi-experts decomposition and nonlinear optimization.
    \item We introduce an AFNO-Conv dual-domain operator that simultaneously captures global cyclical patterns in the frequency domain and transient local microstructures in the temporal domain, addressing the multi-scale nature of financial data. 
    \item Our proposed model, MoFE, achieves state-of-the-art (SOTA) performance across statistical metrics (RMSE, $R^2$) and financial indicators (Sharpe Ratio, Information Coefficient, ROI). High-fidelity back-testing demonstrates significant risk-adjusted returns and robust predictive consistency. 
\end{itemize}

\section{Related Work}

Bitcoin, as the pioneer of decentralized cryptocurrencies, has become one of the most volatile and scrutinized assets in global financial markets~\cite{amjad2017trading}. Unlike traditional financial instruments such as stocks and bonds, Bitcoin operates independently of centralized monetary authorities, and its valuation is influenced by a complex interplay of factors including market liquidity, mining costs, macroeconomic changes, and investor sentiment~\cite{hou2020pricing}. This extreme volatility, structural instability, and inherent non-linearity make Bitcoin price prediction exceptionally difficult. Therefore, establishing a robust forecasting framework is crucial for improving market transparency and providing institutional investors and regulators with a rigorous, data-driven basis for decision-making.

The corpus of existing literature on cryptocurrency forecasting is generally taxonomized into three distinct paradigms. First, statistical methodologies predicate their predictions on linear dynamical systems and stochastic processes—exemplified by ARIMA \cite{tian2023bitcoin} and GARCH \cite{chu2017garch}. These models operate under strong assumptions of stationarity and historical autocorrelation. However, such approaches are inherently constrained to capturing local or univariate features. Consequently, these conventional architectures often fail to simultaneously resolve global frequency-domain patterns and localized temporal dynamics in highly volatile markets. Second, fundamental and sentiment analysis paradigms attempt to quantify intrinsic asset value through on-chain network adoption metrics (e.g., stock-to-flow models) and behavioral indicators~\cite{zargar2019informational}. These methods are premised on the hypothesis that information dissemination and social media discourse serve as leading indicators for subsequent price fluctuations. 

In recent years, deep learning methodologies have established a new state-of-the-art by leveraging non-linear optimization to capture high-dimensional dependencies within financial datasets. Recurrent Neural Network (RNN) variants, most notably Long Short-Term Memory (LSTM)~\cite{seabe2023forecasting} and Gated Recurrent Units (GRU)~\cite{Dutta2020gru}, employ sophisticated gating mechanisms to characterize temporal dynamics. Specifically, the GRU architecture enhances computational efficiency by consolidating the LSTM's forget and input gates into a single update gate, often facilitating superior convergence in Bitcoin volatility modeling. Beyond recurrent paradigms, gradient-boosted frameworks such as XGBoost~\cite{Sekhar2022XGBoost} have demonstrated competitive performance in minimizing mean deviation errors relative to traditional LSTMs. More recently, self-attention-based Transformer architectures~\cite{liu2024itransformer, Kehinde2025Helformer} have achieved significant performance gains by exploiting global long-range dependencies. Despite these advancements, a fundamental limitation still persists: these predominantly time-domain models—including LSTMs and Transformers—exhibit high susceptibility to high-frequency noise, which frequently obscures the latent low-frequency trends inherent in cryptocurrency markets~\cite{peik2025adaptive}. Furthermore, recent studies highlight that models relying exclusively on conventional point-wise regression loss functions are prone to structural phase lag, thereby failing to capture regime-specific dynamics under the extreme non-stationarity of digital assets~\cite{jamhamed2024regime, kang2025harnessing}. While contemporary innovations like Smamba~\cite{gu2023mamba} and CryptoMamba~\cite{Sepehri2025CryptoMamba} attempt to mitigate these issues through enhanced structural generalizability, the inability to disentangle complex dynamics within the frequency domain remains a critical barrier to achieving robust generalization. However, they often struggle with multi-scale feature extraction in the presence of non-stationary volatility.

Frequency-domain methodologies, such as Fourier analysis~\cite{Donghwan2021Fourier} and Fourier Neural Networks~\cite{Runze2025Fourier}, have demonstrated significant efficacy in time-series forecasting by leveraging the inherent time-frequency duality of spectral representations. Concurrently, Mixture-of-Experts frameworks~\cite{Remlinger2023MoE, ni2024mixture, Xiaoming2025TimeMoE} have advanced financial predictive modeling through the dynamic coordination of specialized experts. While recent hybrid architectures like FreqMoE~\cite{Chen2025freqmoe} have successfully integrated FNOs with MoE for solving complex PDE systems, the synergistic application of MoE-driven expert collaboration and FNO-based function-to-function mapping remains largely unexplored in the context of cryptocurrency markets. Consequently, there is a notable research gap in utilizing these combined paradigms to address the pervasive challenges of non-stationarity, abrupt regime shifts, and multi-scale stochasticity inherent in digital asset dynamics.

\section{Methodology}\label{method}

\subsection{Architecture Overview}\label{Architecture}

\begin{figure*}[ht] 
    \centering
    \includegraphics[width=\textwidth]{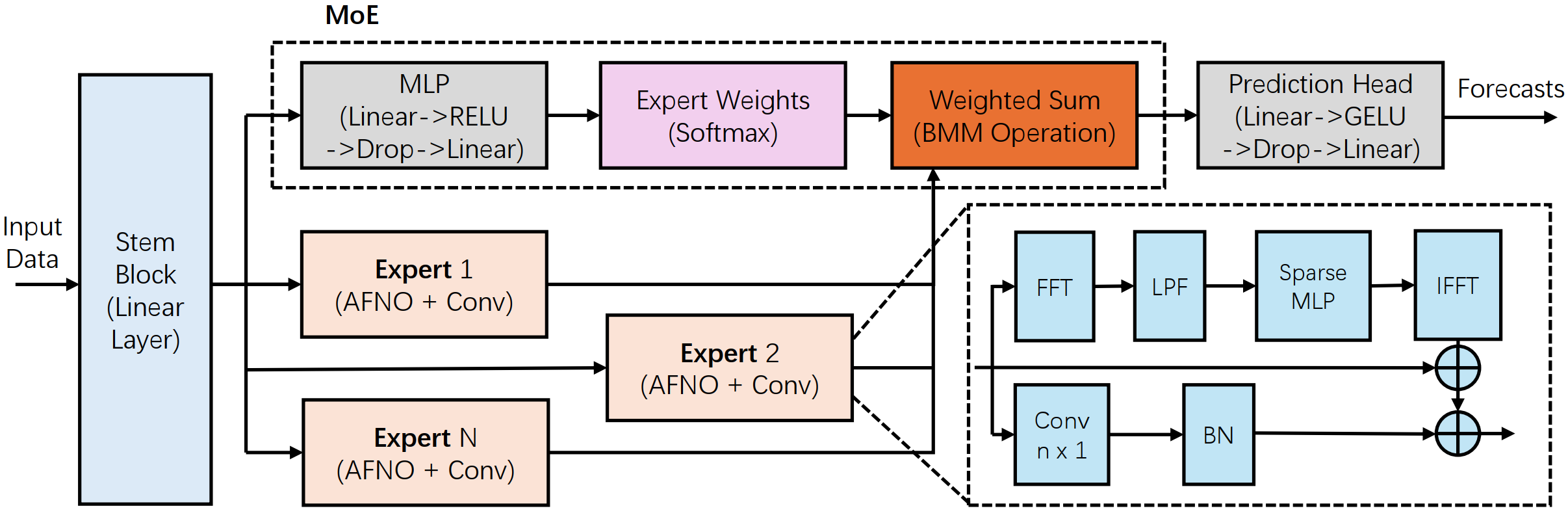} 
    \caption{The overall architecture of the MoFE model. The input features are processed through parallel global (FNO) and local (Conv) paths within the experts, dynamically weighted by the gating network.}
    \label{fig:model_arch}
\end{figure*}

The architecture of the proposed MoFE model is illustrated in Figure~\ref{fig:model_arch}. The process begins with a Stem Block comprising a $\operatorname{Linear}$ layer, which projects the input data sequence into a high-dimensional latent space with $d_{model}$. Subsequently, these features are concurrently processed by $K$ parallel Experts and a Gating Network (MoE block). Each expert is designed as a time-frequency dual-domain feature extraction module. Specifically, the global FNO path employs an adaptive FNO to capture global cyclical fluctuations features via a sparse MLP in the frequency domain, while the local context path utilizes a $n\times1$ convolution (e.g., $n=3$) to extract local microstructure features in the  temporal domain. Simultaneously, the Gating Network regresses expert weights through an MLP and a Softmax layer. These weights are then applied to the expert outputs via Batch Matrix Multiplication (BMM) to dynamically re-weight their contributions. Finally, the aggregated features are passed through a Prediction Head—consisting of $\operatorname{Linear}$, $\operatorname{GELU}$, and $\operatorname{Dropout}$ layers—to produce the final forecasts for prices and returns.

\subsection{Our MoFE Model}\label{Model}
\subsubsection{FNO-Expert}
Each expert $\operatorname{E_k}(\mathbf{X})\in \mathbb{R} ^ {B \times K \times C}$ in our MoFE framework, where $\mathbf{X} \in \mathbb{R} ^ {B \times L \times C}$ denotes the hidden representation (with $B, L, C$ representing batch size, sequence length, and channel dimension $d_{model}$, respectively), employs a dual-path architecture. This design is specifically engineered to synergistically capture global periodic patterns and local transient microstructures. 

\textbf{Global FNO Path - }The core of the FNO-expert utilizes a one-dimensional Adaptive Fourier Neural Operator (AFNO)~\cite{Guibas2022afno} to model long-range dependencies and cyclical market fluctuations. This path operates under the premise that financial time-series evolution can be decomposed into complex modal interactions within the frequency domain. The global path is formulated as:
\begin{equation}
\mathbf{X}_{global} = \operatorname{LN}\left(\mathbf{X} + \mathcal{F}^{-1}\left( \text{LPF} \left( \text{MLP}_{\text{complex}}(\mathcal{F}(\mathbf{X})) \right) \right)\right),
\end{equation}
where $\mathbf{X}_{global}\in \mathbb{R} ^ {B \times K \times C}$, $\mathcal{F}(\cdot)$ is a Fast Fourier Transform (FFT), $\mathcal{F}^{-1}(\cdot)$ is an Inverse FFT (IFFT), $\operatorname{LN}$ is the LayerNorm, $\operatorname{\text{MLP}_{\text{complex}}(\cdot)}$ is a complex-valued MLP block.

The operator $\text{LPF}$ represents a spectral filtering mechanism using a Low Pass Filter (LPF) to mitigate noise and redundancy through two operators: \textit{Hard Thresholding}, which truncates high-frequency components to retain only the most significant low-frequency modes and \textit{Soft Shrinkage ($\operatorname{SS}$)}, which sparsifies the spectral coefficients by filtering out low-amplitude noise with $\operatorname{SS}(\mathbf{Z}) = \operatorname{sign}(\mathbf{Z})\max(0, |\mathbf{Z}| - \lambda)$, where $\lambda$ is a sparsity threshold.

\textbf{Local Context Path - }To complement the spectral path's inherent lack of spatial localization, we introduce a parallel local context branch. It consists of a 1D-convolution with a kernel size of $n$ ($n = 3$), followed by Batch Normalization ($\operatorname{BN}$), to capture short-term momentum, abrupt shocks, and localized trend variations:
\begin{equation} 
\mathbf{X}_{local} = \operatorname{BN}(\text{Conv}{1D}(\mathbf{X})) \end{equation}
where $\mathbf{X}_{local}\in \mathbb{R} ^ {B \times K \times C}$.

\textbf{Feature Fusion - }The global spectral features and local spatial contexts are integrated via element-wise summation. The fused representation is then projected back to the model dimension using a point-wise convolution ($1\times1$) and a $\operatorname{LeakyReLU}$ activation:
\begin{equation}
\operatorname{E_k}(\mathbf{X}) = \text{Conv}(\operatorname{LeakyReLU}(\mathbf{X}_{global} + \mathbf{X}_{local})).
\end{equation}

\subsubsection{Mixture-of-Experts System}

Cryptocurrency markets are characterized by high non-stationarity, exhibiting distinct dynamic regimes (e.g., bullish, bearish, or sideways). The MoE module adaptively manages this regime-switching via a Gating Network, which performs conditional computation by selecting the most relevant experts for a given market state.
The gating mechanism and expert aggregation are defined as:
\begin{equation}
\label{equ:MoE}
    \hat{\mathbf{X}} = \sum_{k=1}^{K} \underbrace{\text{Softmax}(\text{MLP}(\text{vec}(\mathbf{X})))_k}_{\text{Gating weight } g_k} \cdot \underbrace{\operatorname{E}_k(\mathbf{X})}_{\text{Expert output}},
\end{equation}
where $\operatorname{vec}(\cdot)$ is used to flatten the input to a vector, $\operatorname{Softmax}(\cdot)$ is the Softmax operation, the $\operatorname{MLP}(\cdot)$ denotes a learnable MLP that maps the flattened input $\operatorname{vec}(\mathbf{X})\in \mathbb{R} ^ {B \times LC}$ to gating logits. The aggregation is implemented as a batch-wise weighted sum, allowing the model to dynamically reconfigure its internal logic based on the latent state of the batch.

\subsubsection{Prediction Head}
The final forecast is decoded from the aggregated latent features $\hat{\mathbf{X}}$ through a multi-layer perceptron (MLP) based prediction head. To ensure robust mapping and mitigate potential overfitting, the head incorporates non-linear activations and stochastic regularization:
\begin{equation}
\mathbf{Y}_{pred} = \operatorname{Linear}(\text{Dropout}(\text{GELU}(\operatorname{Linear}(\hat{\mathbf{X}})))),
\end{equation}
where $\mathbf{Y}_{pred}\in \mathbb{R} ^ {B \times Horizon}$, $\operatorname{GELU}$ denotes the activation function. This head structure allows the model to refine the high-dimensional latent representations into precise price or return forecasts while maintaining generalization capability.

\subsection{Loss Function}

To address the inherent complexities of financial time-series forecasting, we propose a multi-objective function termed the Break-even Optimized Loss. Unlike conventional approaches that rely solely on Euclidean distance minimization, our composite loss functional integrates magnitude accuracy, correlation-based alignment, and directional consistency to better capture market dynamics:
\begin{equation} 
\mathcal{L}_{Task} = \alpha \mathcal{L}_{MSE}^{w} + \beta \mathcal{L}_{IC} + (1-\alpha-\beta) \mathcal{L}_{Dir}, 
\end{equation}
where the components$\mathcal{L}_{MSE}^{w}$, $\mathcal{L}_{IC}$, and $\mathcal{L}_{Dir}$ represent the weighted Mean Squared Error of log return, Information Coefficient loss, and Directional Accuracy loss, respectively. These metrics are grounded in established financial econometrics and forecasting literature \cite{grinold1999active, diebold1996forecast, pesaran1992simple}. The hyperparameters $\alpha$, $\beta$ govern the trade-offs between these competing objectives.

From a mathematical standpoint, discrete post-hoc strategy metrics derived from executed trading signals, such as the Win/Loss ratio, are non-differentiable. Therefore, they are structurally incompatible as direct penalty terms in backpropagation.

However, the proposed Break-even Optimized Loss is explicitly architected to achieve this trade-off without compromising differentiability via the existing continuous hyperparameters $\alpha$ and $\beta$. Specifically, Magnitude Accuracy is directly governed by $\alpha$ (which weights the LMSE term), whereas Directional Frequency is controlled by the residual weight $(1-\alpha-\beta)$ for LDir and $\beta$ for LIC.

This formulation allows practitioners to tune the existing $\alpha$ and $\beta$ parameters to customize the model's risk-reward profile. For instance, aggressive trading strategies prioritizing directional hit rates (Win/Loss) over magnitude suppression can simply lower the $\alpha$ weight, thereby achieving trading style flexibility.

A common challenge in MoE architectures is ``expert collapse" (or mode collapse), where the gating mechanism converges to a state that activates the same subset of experts for all inputs, thereby sacrificing model capacity. To enforce expert specialization and diversity, we introduce an orthogonality constrained regularization loss $\mathcal{L}_{Reg}$ on the gating activations,
\begin{equation} 
\mathcal{L}_{Total} = \mathcal{L}_{Task} + \gamma \mathcal{L}_{Reg}, 
\end{equation}
where $\gamma$ is the weight of the regularization loss.

Let $\mathbf{G} \in \mathbb{R}^{B \times K}$ denote the gating activation matrix for a batch size $B$ with $K$ experts. We first perform $L_2$ normalization on the activation vectors $g_k$ ($g_k  \in \mathbb{R}^{B \times 1}$, seen in Equ.~\ref{equ:MoE}) of each expert $k$ (representing the $k$-th column of $\mathbf{G}$) to ensure scale invariance:
\begin{equation} 
\hat{g}_k = \frac{g_k}{|g_k|_2 + \epsilon}, 
\end{equation}
where $|\cdot|_2$ is an $L_2$ normalization, $\epsilon$ is a small constant for numerical stability, resulting in the normalized matrix $\hat{\mathbf{G}} \in \mathbb{R}^{B \times K}$.

To promote decorrelation among experts, we penalize the off-diagonal elements of the Gram matrix $\hat{\mathbf{G}}^T \hat{\mathbf{G}} \ \in \mathbb{R}^{K \times K}$ to drive it toward the identity matrix $\mathbf{I}$ as
\begin{equation}
\mathcal{L}_{Reg} = \| \hat{\mathbf{G}}^T \hat{\mathbf{G}} - \mathbf{I} \|_F,
\end{equation}
where $\|\cdot\|_F$ denotes the Frobenius norm. This regularization term effectively encourages disjoint expert activation patterns, ensuring that different experts specialize in distinct market regimes or latent feature subspaces. 

\section{Experimental Analysis and Discussion}\label{experiment}

\subsection{Datasets}

The dataset for this study was collected via a bespoke web crawler from historical Bitcoin repositories, spanning the period from January 1, 2020 to December 27, 2025. The raw data comprises daily Open-High-Low-Close-Volume (OHLCV) metrics organized in strict chronological order. To maintain temporal integrity, the dataset was physically partitioned into training, validation, and testing sets with an approximate ratio of 4:1:1 prior to any preprocessing.

The dataset for this study was collected using a Python script accessing the Binance Public API (specifically the v3/klines endpoint). The raw exchange data comprises daily cryptocurrency K-line data organized in strict chronological order, spanning the period from January 1, 2020, to December 27, 2025. It includes six primary attributes: Timestamp, Open, High, Low, Close, and Volume.

To preclude data leakage, normalization was applied locally to each sliced window, while global standardization parameters and z-score were derived exclusively from the training set. All data used in the test set, including for forecasting inference and performance metric calculations, only uses data from before the given time point, and not data from after that time point.

It is imperative to explicitly recognize the scope and boundaries of our methodology. First, our data collection strictly relies on endogenous market variables (OHLCV data) from the Binance exchange, explicitly excluding external macroeconomic indicators (e.g., interest rates, inflation) that might drive long-term valuations. Second, while the MoFE model demonstrates robustness, its predictive reliability may diminish under extreme market conditions or systemic "black swan" events where historical patterns fundamentally break down. Finally, the current 32-day look-back window may be insufficient to capture very low-frequency cyclical patterns, such as the quadrennial Bitcoin halving events.

To enhance the representative capacity of the MoFE model, the feature space is expanded into an 8-dimensional vector. This includes the logarithmic returns of the OHLC prices, supplemented by four technical indicators, including Relative Strength Index (RSI), Moving Average Convergence Divergence (MACD), Relative Volume (RVol), and a Volatility Proxy (\textit{VolProxy}). The \textit{VolProxy} characterizes price fluctuations based on the log returns over a rolling window of $n=5$ days. These features are derived according to established technical analysis methodologies \cite{pring2002technical, appel2005technical} to capture both momentum and volatility dynamics.

\subsection{Performance Metrics}

To rigorously evaluate the predictive efficacy and practical profitability of the MoFE model within the volatile Bitcoin market, we implement a multi-dimensional evaluation framework. This framework is categorized into three dimensions: Statistical Predictive Metrics, Trading Strategy Metrics, and Profitability Metrics.

Statistical forecasting metrics quantify a model’s ability to minimize forecast error and capture market return variance. These metrics include root mean square error (RMSE), mean absolute error (MAE), and $R^2$~\cite{hyndman2006another}. These metrics are all back-calculated prices statistical error computed from the log return forecasting from the MoFE. The trading strategy metrics are used to evaluate the quality and persistence of investment signals or ``alpha models”, including
Information Coefficient (IC)~\cite{grinold1999active} and Directional Accuracy (DA)~\cite{pesaran1992simple}. 
Beyond statistical accuracy, we assess the model's utility in simulated trading environments, including Win/Loss Ratio~\cite{pardo2008evaluation}, Cumulative Return (ROI)~\cite{grinold1999active}, and Annualized Sharpe Ratio~\cite{sharpe1994sharpe}. In this experiment, Transaction Fee rate and slippage fee are set $0.1\%$ and $0.2\%$ respectively on day-line trading.

\subsection{Experiments Setup}

All proposed models and baselines were implemented using the PyTorch framework and executed on a platform equipped with an NVIDIA GeForce RTX 4060 Laptop GPU. To ensure experimental rigor and a fair benchmarking environment, we maintained consistent configurations across all evaluated models. This includes a fixed random seed ($42$) for deterministic behavior, uniform data partitioning, identical input features.

All models were optimized using the AdamW optimizer with a standardized learning rate schedule.

All models were optimized using the AdamW optimizer. To mitigate overfitting and ensure optimal generalization, we incorporated a ReduceLROnPlateau learning rate scheduler, weight decay, and an early stopping mechanism with a patience of 20 epochs. The model checkpoint with the lowest validation loss was selected for testing.

Tab.~\ref{tab:hyperparams} summarizes the experimental configuration. We utilize a look-back window of $L=32$ days for forecasting cumulative log-returns. Given the high susceptibility of $T+1$ predictions to transient noise, our evaluation emphasizes the $T+5$ horizon to better assess the capture of persistent temporal dependencies. The model dimension is set to $d_{model}=48$. Within the MoE-FNO framework, we deploy $K=2$ experts to prevent overfitting, with each expert operating on $M=8$ frequency modes. This design strikes an optimal balance between model capacity and computational efficiency.

\begin{table}[htbp]
    \centering
    \caption{Hyperparameter Configuration of MoFE}
    \label{tab:hyperparams}
    \begin{tabular}{l l c}
        \toprule
        \textbf{Category} & \textbf{Hyperparameter} & \textbf{Value} \\
        \midrule
        \textbf{Data} & Look-back Window ($L$) & 32 \\
                      & Forecast Horizon ($T_{p}$) & 1, 5 \\
        \midrule
        \textbf{Model} & Hidden Dimension ($d_{model}$) & 48 \\
                       & Frequency Modes & 8 \\
                       & Number of Experts ($K$) & 2 \\
                       & FNO Blocks & 1 \\
                       & Dropout Rate & 0.5 \\
        \midrule
        \textbf{Training} & Batch Size & 128 \\
                          & Initial Learning Rate & $5 \times 10^{-4}$ \\
                          & Weight Decay & $1 \times 10^{-2}$ \\
                          & Loss Weights ($\alpha, \beta$) & 0.20, 0.30 \\
                          & Reg. Loss Weights ($\gamma$) & 0.05 \\
        \bottomrule
    \end{tabular}
\end{table}

\subsection{Price Forecasting Experiments}\label{AAA}

\subsubsection{Overall Performance}

The quantitative results of our comparative study are summarized in Tab.~\ref{tab:main_results1} and \ref{tab:main_results2}. We benchmark the proposed MoFE framework against a suite of state-of-the-art baselines across prediction horizons of $T+1$ and $T+5$. 

In terms of specific comparisons, traditional RNN-based architectures (e.g., LSTM\cite{seabe2023forecasting} and GRU\cite{Dutta2020gru}) achieve competitive DA but yield suboptimal results in profitability and precision compared to MoFE. While the iTransformer\cite{liu2024itransformer} demonstrates competence in select metrics, it generally lacks the comprehensive competitiveness of our proposed method. Furthermore, although recent state-space models such as CryptoMamba\cite{Sepehri2025CryptoMamba} exhibit promising potential, MoFE maintains a clear performance margin. Moreover, other MoE models like FreqMoE\cite{Chen2025freqmoe} and  MoLE\cite{ni2024mixture} are also overshadowed by MoFE. The empirical evidence indicates that MoFE consistently surpasses existing methods, demonstrating significant improvements in both statistical accuracy and economic utility. This superiority is largely attributed to MoFE's modeling capability on cryptocurrency non-stationary volatilization.

\begin{table*}[htbp]
  \centering
  \caption{Performance comparison of MoFE against baseline models for T+1 prediction. The best results are in \textbf{bold}.}
  \label{tab:main_results1}
  \renewcommand{\arraystretch}{1.2}
  \setlength{\tabcolsep}{4pt}

  \begin{tabular}{lcccccccc}
    \toprule
    \multirow{2}{*}{\textbf{Model}} & \multicolumn{3}{c}{\textbf{Statistical Metrics}} & \multicolumn{2}{c}{\textbf{Strategy Metrics}} & \multicolumn{3}{c}{\textbf{Profitability Metrics}} \\
    
    \cmidrule(lr){2-4} \cmidrule(lr){5-6} \cmidrule(lr){7-9}
     & RMSE/\$ ($\downarrow$) & MAE/\$ ($\downarrow$) & $R^2$ ($\uparrow$) & IC ($\uparrow$) & DA (\%) ($\uparrow$) & Sharpe ($\uparrow$) & W/L Ratio ($\uparrow$) & ROI (x) ($\uparrow$) \\
    \midrule
    LSTM\cite{seabe2023forecasting} & 1618.35 & 1134.32 & 0.9819 & 0.0611  & 54.52\% & 1.00 & 1.00 & 1.48 \\
    GRU\cite{Dutta2020gru} & 1605.20 & 1130.03 & 0.9819 & 0.0725 & 53.01\% & 1.03 & \textbf{1.24} & 1.16 \\
    iTransformer\cite{liu2024itransformer} & 1603.96 & 1130.63 & 0.9822 & 0.1225 & 51.51\% & 2.42 & 0.96 & 2.75 \\
    S-Mamba\cite{gu2023mamba} & 1662.95 & 1183.44 & 0.9809 & 0.0127  & 54.82\% & 1.40 & 0.96 & 1.86 \\
    CryptoMamba\cite{Sepehri2025CryptoMamba} & 1613.97 & 1145.64 & 0.9820 & 0.1123 & 51.81\% & 2.93 & 1.17 & 3.13 \\
    FreqMoE\cite{Chen2025freqmoe} & 1607.94 & 1140.63 & 0.9821 & 0.1156 & 52.41\% & 1.34 & 1.09 & 1.69 \\
    MoLE\cite{ni2024mixture} & 1621.76 & 1151.95 & 0.9818 & 0.1181 & 55.72\% & 2.11 & 1.07 & 2.45 \\
    \midrule
    \textbf{MoFE (Ours)} & \textbf{1582.64} & \textbf{1106.10} & \textbf{0.9827} & \textbf{0.1957} & \textbf{58.43\%} & \textbf{3.89} & 1.09 & \textbf{4.41} \\
    \bottomrule
  \end{tabular}
\end{table*}

\begin{table*}[htbp]
  \centering
  \caption{Performance comparison of MoFE against baseline models for T+5 prediction. The best results are in \textbf{bold}.}
  \label{tab:main_results2}
  \renewcommand{\arraystretch}{1.2}
  \setlength{\tabcolsep}{4pt}
  
  \begin{tabular}{lcccccccc} 
    \toprule
    \multirow{2}{*}{\textbf{Model}} & \multicolumn{3}{c}{\textbf{Statistical Metrics}} & \multicolumn{2}{c}{\textbf{Strategy Metrics}} & \multicolumn{3}{c}{\textbf{Profitability Metrics}} \\
    \cmidrule(lr){2-4} \cmidrule(lr){5-6} \cmidrule(lr){7-9}
     & RMSE/\$ ($\downarrow$) & MAE/\$ ($\downarrow$) & $R^2$ ($\uparrow$) & IC ($\uparrow$) & DA (\%) ($\uparrow$) & Sharpe ($\uparrow$) & W/L Ratio ($\uparrow$) & ROI (x) ($\uparrow$) \\
    \midrule
    LSTM\cite{seabe2023forecasting} & 3392.81 & 2596.66 & 0.9170 & 0.1015 & \textbf{56.10\%} & 1.07 & 0.68 & 1.07 \\
    GRU\cite{Dutta2020gru} & 3410.93 & 2618.38 & 0.9146 & 0.0326 & 52.44\% & 1.11 & 1.15 & 0.90 \\
    iTransformer\cite{liu2024itransformer} & 3381.62 & \textbf{2592.32} & 0.9176 & 0.0760 & 53.66\% & 1.07 & \textbf{1.37} & 1.50 \\
    S-Mamba\cite{gu2023mamba} & 3443.05 & 2620.90 & 0.9146 & 0.0259  & 54.88\% & 0.82 & 1.05 & 0.92 \\
    CryptoMamba\cite{Sepehri2025CryptoMamba} & 3650.63 & 2798.99 & 0.9040 & 0.0473 & 54.88\% & 0.96 & 1.17 & 1.60 \\
    FreqMoE\cite{Chen2025freqmoe} & 3414.83 & 2620.41 & 0.9160 & 0.0678 & 55.18\% & 1.29 & 1.22 & 1.82 \\
    MoLE\cite{ni2024mixture} & 3493.51 & 2699.26 & 0.9120 & 0.0142 & 51.83\% & 0.36 & 1.11 & 0.40 \\
    \midrule
    \textbf{MoFE (Ours)} & \textbf{3360.33} & 2593.37 & \textbf{0.9186} & \textbf{0.1389} & \textbf{56.10\%} & \textbf{1.73} & 1.26 & \textbf{4.91} \\
    \bottomrule
  \end{tabular}
\end{table*}

\subsubsection{T+1 Forecasting}
~\\

\begin{figure}[ht] 
    \centering
    \includegraphics[width=\linewidth]{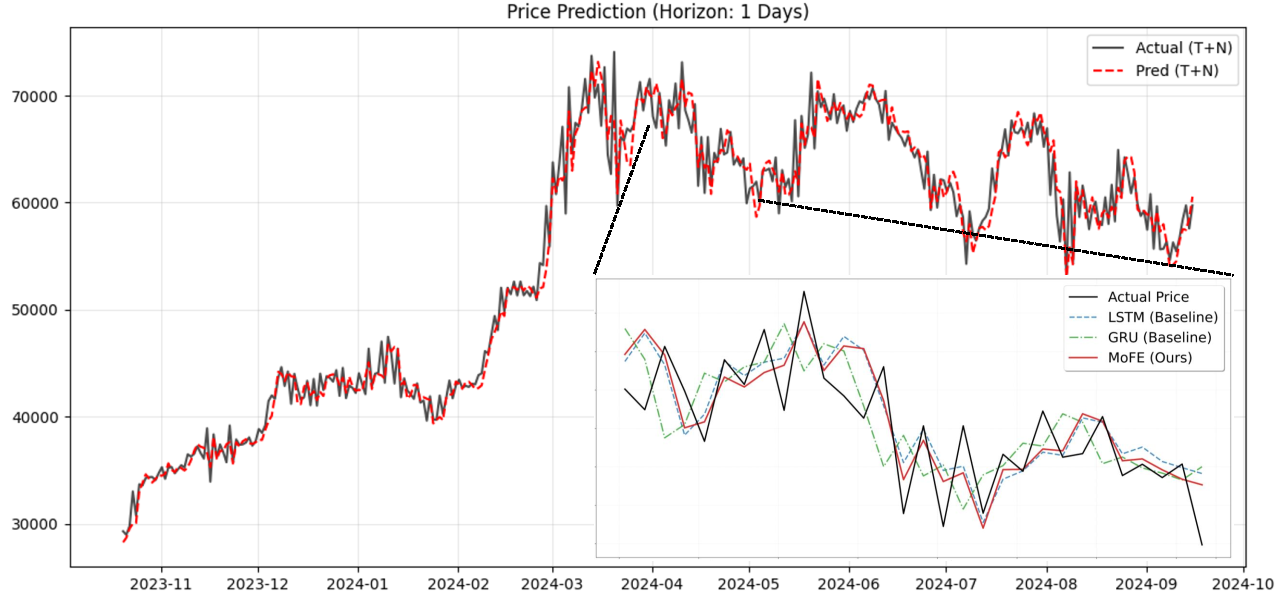} 
    \caption{Visualized results of the comparison of predicted prices and real prices.}
    \label{fig:price_prediction}
\end{figure}

\indent\textbf{Predictive Accuracy Analysis:}
It can be concluded from Tab.~\ref{tab:main_results1} and Fig.~\ref{fig:price_prediction} that 

MoFE outperforms all baselines, yielding a minimum RMSE of \$1582.64 and a maximum $R^2$ of 0.9827, signaling a precise mapping of continuous market dynamics. Unlike conventional models that often suffer from significant phase lag by collapsing into a trivial identity function ($P_{t+1} \approx P_t$), MoFE captures genuine structural shifts. This distinction is vital in finance, where high autocorrelation often inflates RMSE and $R^2$ scores through a naive Lag-1 imitation of price persistence. By maintaining high DA far above the 50\% baseline, MoFE proves its robustness against the deceptive optimism of standard statistical metrics and its ability to identify actionable volatility clusters.

\textbf{Directional Predictive Capability:}
Beyond point-wise convergence, MoFE demonstrates superior directional predictive strength. From Tab.~\ref{tab:main_results1}, at the $T+1$ horizon, it achieves a peak IC of 0.1957 and DA of 58.43\%, with its IC exceeding the runner-up baseline by 1.6$\times$. This suggests an exceptionally robust monotonic correspondence between predictive signals and realized returns.

\begin{figure}[ht] 
    \centering
    \includegraphics[width=\linewidth]{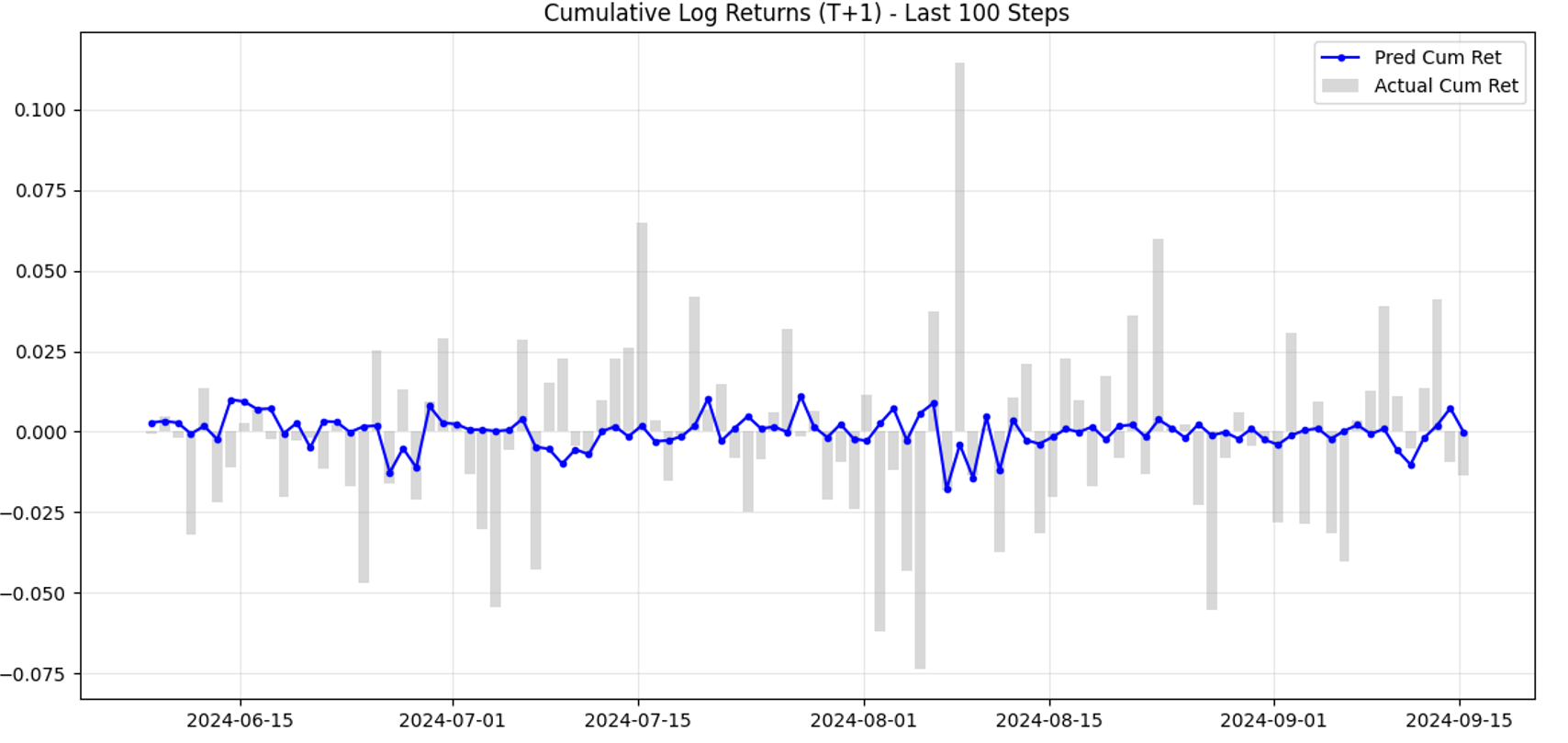}     
    \caption{Visualized results of the log returns prediction versus the real situations.}
    \label{fig:cumulative_log_return}
\end{figure}

\begin{figure}[ht] 
    \centering
    \includegraphics[width=\linewidth]{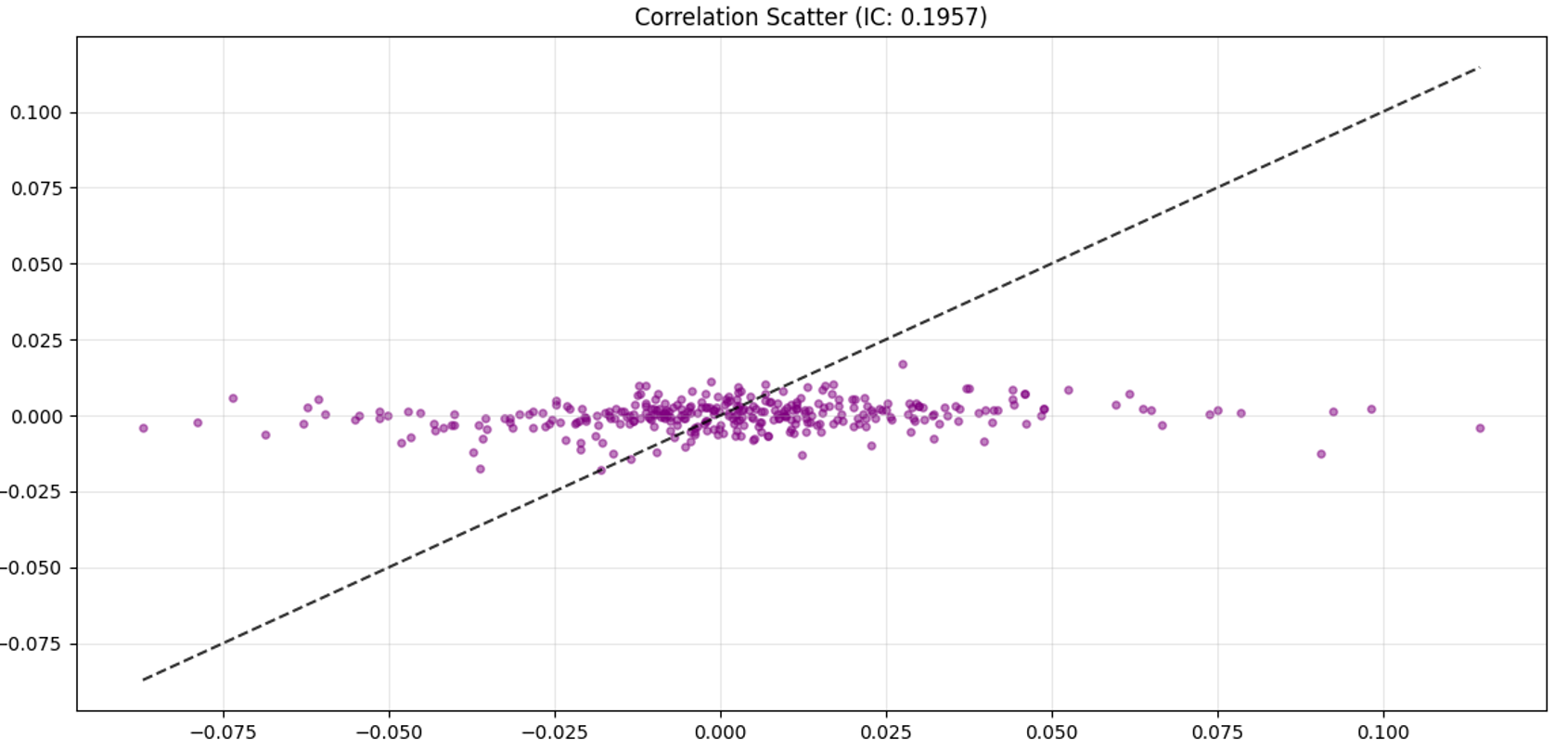} 
    \caption{Visualized results of the correlation scatter of the predicted log-returns and the realized market returns.}
    \label{fig:correlation_scatter}
\end{figure}

Visualized results in Figs. \ref{fig:cumulative_log_return} and \ref{fig:correlation_scatter} further elucidate these findings below: (1) In Fig. \ref{fig:cumulative_log_return}, the predicted cumulative returns (blue line) exhibit a dampened amplitude relative to market volatility (grey bars). This reduced variance underscores the efficacy of our balanced loss function, which induces strategic conservatism by penalizing significant directional deviations. (2) Despite this magnitude suppression, the predicted trajectory closely tracks actual market trends. The scatter plot in Fig.~\ref{fig:correlation_scatter} confirms this, with the points densely clustered around the diagonal line ($y=x$), indicating a strong linear correlation. Simultaneously, the clustering of points near $y=0$ reflects the  risk-aversion effect of the loss function. These results collectively provide empirical evidence that MoFE can effectively filter out high-frequency noises while preserving high-fidelity directional signals, which is consistent with its superior IC and DA metrics.

\textbf{Financial Performance Evaluation:}
MoFE model achieved the superior performance in terms of ROI (4.41) and Sharpe Ratio (3.89), which indicates that the model not only achieves a superior return rate while maintaining a low risk profile. This shows that MoFE model yields substantial excess returns in unit risk.
Although MoFE has Win/Loss Ratio a slightly lower than GRU and iTransformer, it still performs better in overall return and risk control.

\subsubsection{T+5 Forecasting}
~\\
\indent\textbf{Predictive Accuracy Analysis:}
Even though having slightly higher MAE (\$2593.37) than iTransformer model only at the difference of 1.05, MoFE model still has the smallest RMSE (\$3360.33) and $R^2$ (0.9186) in $T+5$ price prediction. Since RMSE is more sensitive to large errors, the results showed that the MoFE model is more stable.

\textbf{Directional Predictive Capability:}
Similar to the situation of T+1 prediction, MoFE model also took the lead in IC (0.1389) and DA (56.10\%).

\textbf{Financial Performance Evaluation:}
In $T+5$ prediction, MoFE model also performed best in ROI (4.91) and Sharpe Ratio (1.73). Its exceptional ROI significantly outperforms competing models, surpassing the second-best approach by a factor of three.
Even though it still did not get the highest Win/Loss Ratio. It remains the optimal architecture for maximizing risk-adjusted returns.

\subsubsection{Robustness and Horizon Sensitivity}

While the horizon extends from 1 to 5, all models exhibit a natural deterioration in Predictive Metrics, characterized by increased RMSE/MAE and decreased $R^2$. However, MoFE demonstrates superior robustness, suffering only a marginal and controllable decline. Crucially, its ROI improves rather than declines over this longer horizon, which highlights its ability to grasp mid-term market dynamics. The results indicate that that MoFE model can not only dominate short-term forecasting but also leads in medium-term scenarios.

\subsection{Efficiency Analysis}

The MoFE model demonstrates high computational and parameter efficiency for both $T$+1 and $T$+5 forecasting tasks. Specifically, the model requires 93.6K parameters and consumes 2.36M FLOPs in $T$+1 mode, while the $T$+5 configuration requires 94.1K parameters with identical FLOPs. These metrics are highly competitive with established baseline models.

Crucially, despite integrating a complex dual-domain structure, MoFE achieves SOTA performance with a highly lightweight footprint, making it explicitly competitive against more parameter-heavy baselines such as iTransformer and Mamba-based models, which typically demand significantly higher memory and computational overhead.

\subsection{Interpretability Analysis}\label{Impact}
 
\begin{table*}[htbp]
  \centering
  \caption{Ablation Study}
  \label{tab:Ablation Study}
  \renewcommand{\arraystretch}{1.2}
  \setlength{\tabcolsep}{4pt}      

  \begin{tabular}{lcccccccc}
    \toprule
    \multirow{2}{*}{\textbf{Model}} & \multicolumn{3}{c}{\textbf{Statistical Metrics}} & \multicolumn{2}{c}{\textbf{Strategy Metrics}} & \multicolumn{3}{c}{\textbf{Profitability Metrics}} \\
    \cmidrule(lr){2-4} \cmidrule(lr){5-6} \cmidrule(lr){7-9}
     & RMSE/\$ ($\downarrow$) & MAE/\$ ($\downarrow$) & $R^2$ ($\uparrow$) & IC ($\uparrow$) & DA (\%) ($\uparrow$) & Sharpe ($\uparrow$) & W/L Ratio ($\uparrow$) & ROI (x) ($\uparrow$) \\
    \midrule
    1 Expert (w/o MoE) & 1608.70 & 1127.54 & 0.9821 & 0.0903 & 51.51\% & 1.66 & \textbf{1.37} & 1.43 \\
    4 Experts (w MoE) & 1658.03 & 1184.79 & 0.9810 & 0.0680 & 53.01\% & 1.29 & 1.16 & 1.70 \\
    2 Experts (w/o MoE) & 1633.03 & 1142.99 & 0.9816 & 0.1077 & 54.52\% & 0.67 & 1.07 & 1.76 \\
    Conv. Kernel in Experts ($n=1$) & 1597.61 & 1161.83 & 0.9824 & 0.1627 & 55.12\% & 2.11 & 1.13 & 2.52 \\
    Conv. Kernel in Experts ($n=5$)& 1602.83 & 1129.21 & 0.9823 & 0.1324 & 54.22\% & 3.31 & 1.12 & 2.66 \\
    w/o Reg. Loss & \textbf{1582.59} & 1107.91 & \textbf{0.9827} & 0.1937 & 56.93\% & \textbf{3.91} & 1.07 & \textbf{4.44} \\
    \textbf{Ours} & 1582.64 & \textbf{1106.10} & \textbf{0.9827} & \textbf{0.1957} & \textbf{58.43\%} & 3.89 & 1.09 & 4.41 \\
    \bottomrule
  \end{tabular}
\end{table*}

\begin{figure}[ht] 
    \centering
    \includegraphics[width=\linewidth]{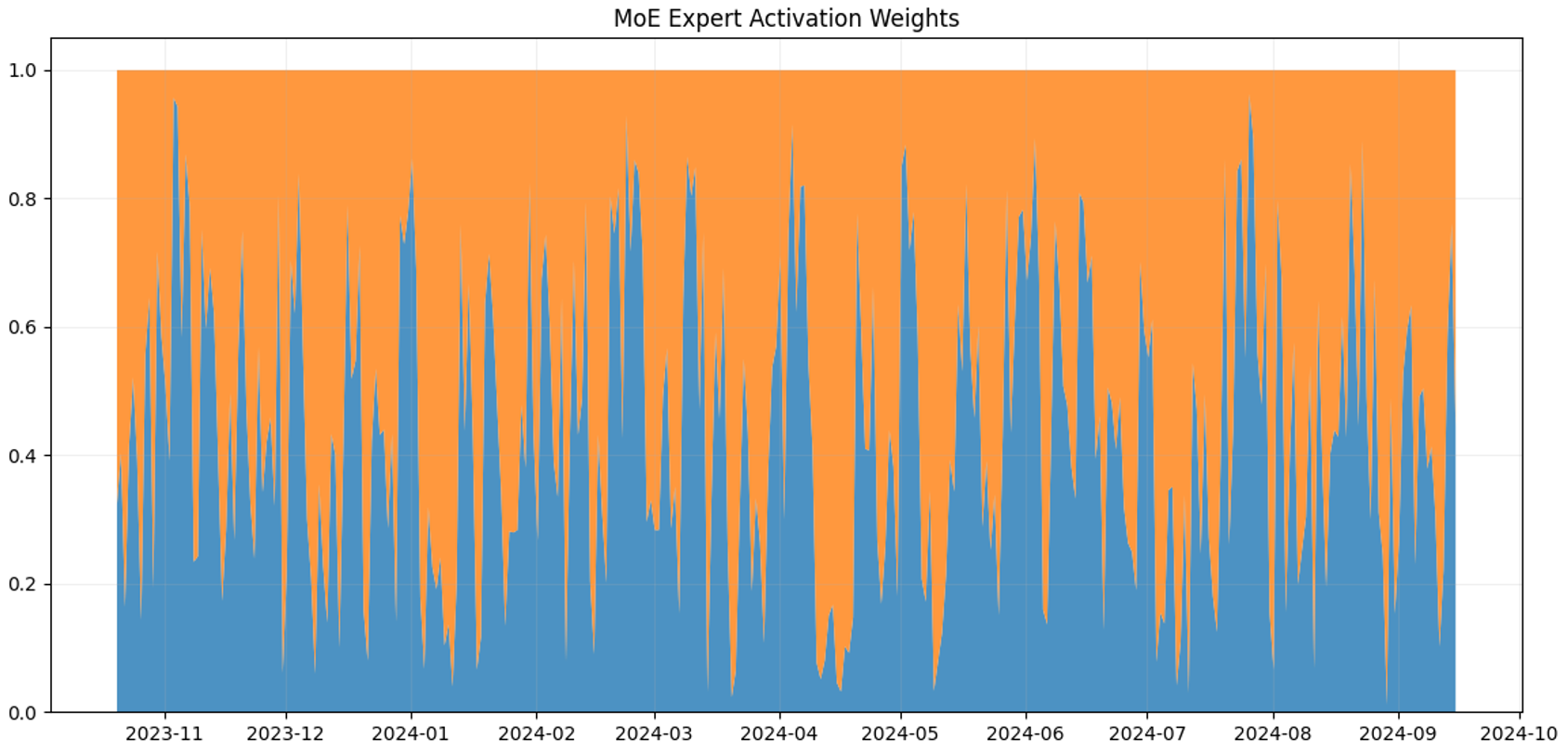} 
    \caption{Visualized results of two experts' dynamic weights during mixing.}
    \label{fig:moe_expert_activation_weights}
\end{figure}

Figure \ref{fig:moe_expert_activation_weights} indicates that the two experts will change the weight according to different conditions. Sometimes, the experts represented in blue dominate the choice depicted in orange gets more weights. By using the Reg. Loss, the two experts are incentivized to capture distinct feature subspaces. One focuses on short-term fluctuations and the other one focus on long-term trend. This enables the model to possess both the time and frequency domain perspective.

\begin{figure}[ht] 
    \centering
    \includegraphics[width=\linewidth]{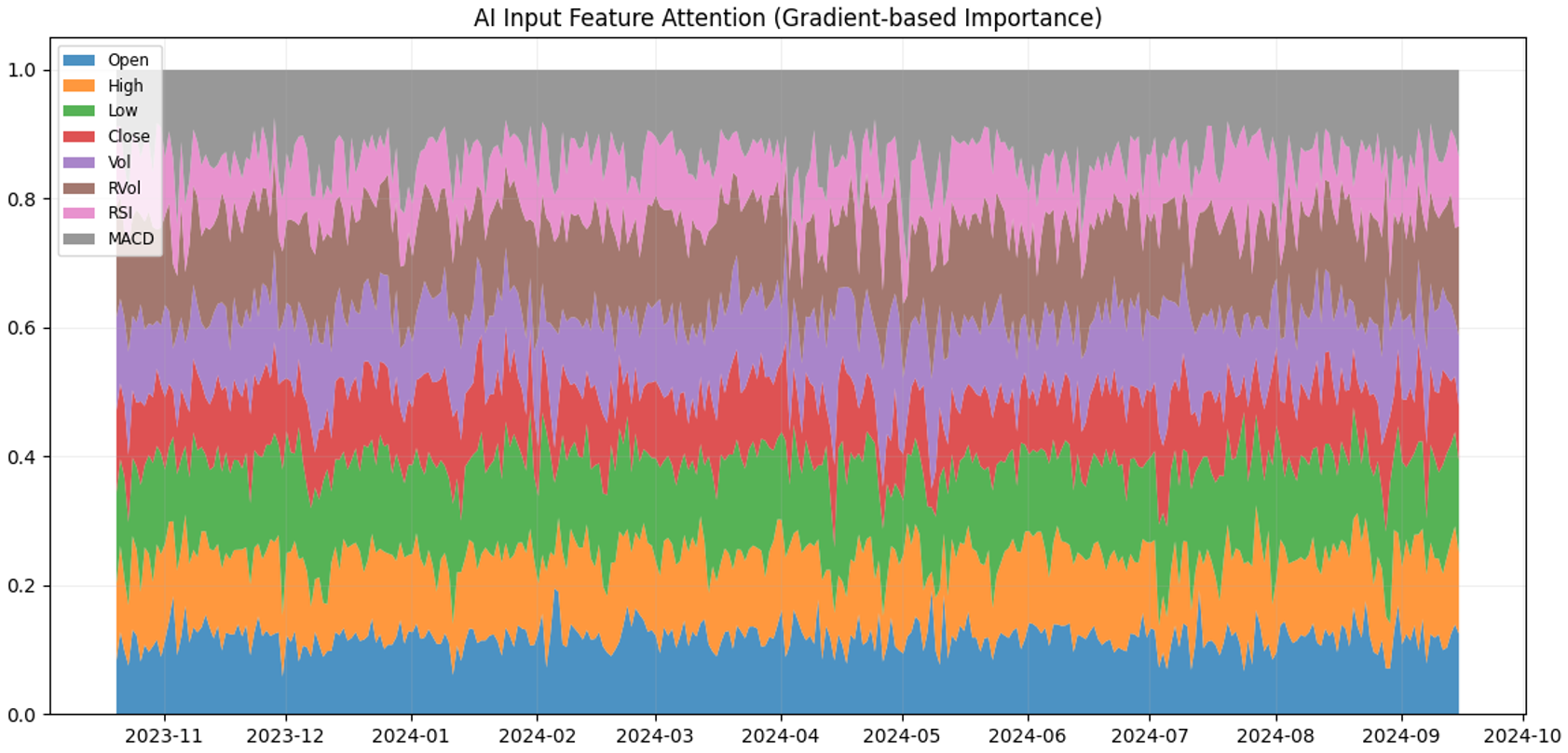} 
    \caption{Visualized results of the input vector contribution for the forecasting.}
    \label{fig:ai_input_feature_attention}
\end{figure}

Figure \ref{fig:ai_input_feature_attention} shows that the weight of the 8 input features varies according to different condition. By adjusting the weight dynamically, the model can adaptively respond to different conditions in the Bitcoin market. Those factors have roughly the same weight in most of the cases, however, when the market experiences high volatility the model will react and recalibrate weights to make the prediction more accurate.

\subsection{Ablation Study}

We conduct a series of ablation experiments to evaluate the contribution of each core component in MoFE. The quantitative results are summarized in Table \ref{tab:Ablation Study}.

1) Impact of regularization loss ($L_{reg}$) - Removing $L_{reg}$ leads to a marked decline in DA and Win/Loss Ratio, validating that $L_{reg}$ effectively encourages expert specialization. While metrics such as Sharpe Ratio and ROI exhibit marginal decreases, the overall enhancement in predictive precision confirms that $L_{reg}$ prevents expert collapse and enables the model to capture diverse market dynamics more accurately.

\begin{figure}[ht] 
    \centering
    \includegraphics[width=\linewidth]{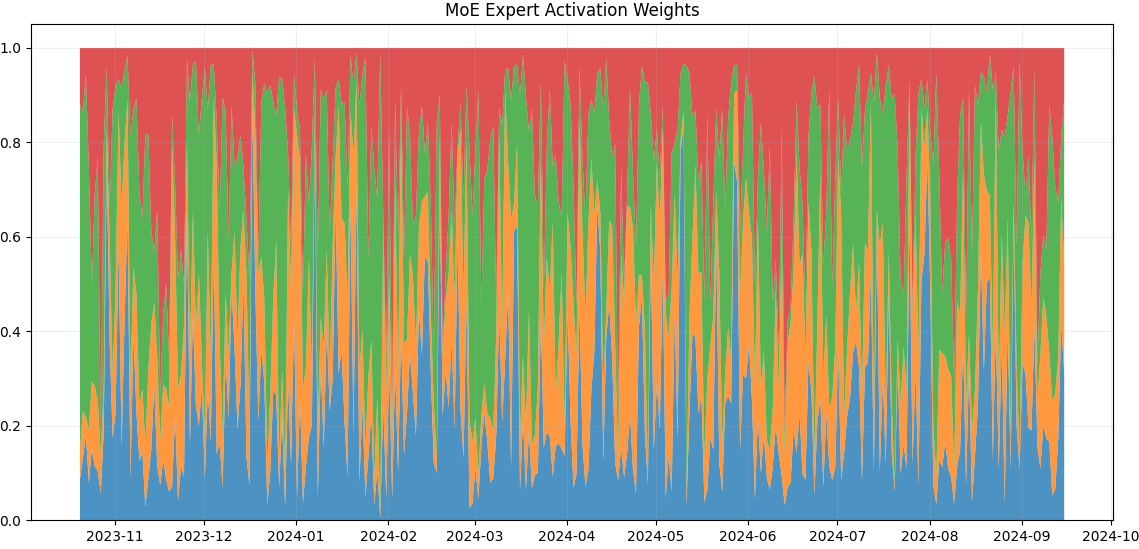}  
    \caption{Visualized results of four experts' dynamic weights during mixing.}
    \label{fig:MoE_Ablation}
\end{figure}

2) Scaling the number of experts ($N$) - Varying the expert count from $N$=2 to $N$=1 or $N$=4 results in performance degradation across most metrics. As illustrated in Fig. \ref{fig:MoE_Ablation}, increasing $N$ to 4 leads to ``expert dilution", where the gating weights fluctuate erratically, suggesting that the 32-day look-back window provides insufficient information to supervise a high-dimensional expert space, thereby inducing overfitting. Conversely, a single expert ($N$=1) fails to simultaneously resolve high-frequency noise and medium-term trends. The $N$=2 configuration achieves the optimal balance by dedicating specialized capacity to distinct temporal scales.

3) Efficacy of the gating mechanism - To isolate the benefit of dynamic routing, we replaced the gating network with a simple averaging operation while retaining $L_{reg}$. The results show a significant deterioration in IC, DA, and profitability metrics. This drop confirms that the gating network is not merely an aggregator but a critical decision-maker that dynamically prioritizes experts based on shifting market regimes.

4) Sensitivity to convolutional kernel size - The spatial receptive field of each expert was tested with kernel sizes of $1\times1$, $3\times1$, and $5\times1$. Reducing the kernel to $1\times1$ severely impairs strategic performance, as the model loses its ability to extract local temporal correlations. However, expanding the kernel to $5\times1$ also degrades performance, likely due to the introduction of excessive noise from distant time steps. A $3\times1$ kernel proves to be the ``sweet spot" for capturing meaningful medium-term trends without overfitting to stochastic fluctuations.

\section{Conclusion}\label{conclusion}
We introduced MoFE in this work, a novel deep learning framework that integrates FNOs and MoE to capture the non-stationary and multi-scale stochasticity of cryptocurrency markets. By leveraging a dual-domain architecture, MoFE effectively extracts global spectral trends and local temporal microstructures, while a dynamic gating mechanism ensures adaptability to rapid regime shifts. 

Our experiments confirm that MoFE achieves SOTA results for $T+1$ and $T+5$ horizons, significantly reducing phase lag and enhancing directional predictive power.

This superior phase-lag mitigation is achieved because the frequency-domain modeling effectively separates high-frequency noise from underlying structural market trends. Empirically, this theoretical advantage translates into robust performance particularly during high-volatility periods, where the spectral representation seamlessly isolates structural trends from transient short-term noise.

In simulated trading, the model yielded exceptional risk-adjusted returns, demonstrating high practical utility. 

Future research will strictly focus on extending the MoFE framework to other financial time-series forecasting tasks, such as cross-asset volatility modeling.

\ifCLASSOPTIONcaptionsoff
  \newpage
\fi
{
	\small
	\bibliographystyle{IEEEtran}
	\bibliography{mybibfile}
}

\end{document}